%% file: main.tex
\documentclass{article}

\PassOptionsToPackage{numbers,sort&compress}{natbib}
\IfFileExists{neurips_2026.sty}{
  \usepackage[preprint]{neurips_2026}
}{
  \usepackage[margin=1in]{geometry}
}

\usepackage[utf8]{inputenc}
\usepackage[T1]{fontenc}
\usepackage{url}

\usepackage{graphicx}
\usepackage{subcaption}
\usepackage{booktabs}
\usepackage{amsfonts}
\usepackage{amsmath,amssymb}
\usepackage[protrusion=true,expansion=false]{microtype}
\usepackage{xcolor}

\usepackage[accsupp]{axessibility}  % Improves PDF readability for those with disabilities.

\usepackage[ruled]{algorithm2e}

\SetKwProg{Fn}{Function}{:}{end}
\SetAlFnt{\small}
\SetAlCapFnt{\small}
\SetAlCapNameFnt{\small}
\SetAlCapHSkip{0pt}
\SetAlgoSkip{4pt}

\usepackage{hyperref}

\usepackage{orcidlink}

\usepackage{multirow}
\usepackage{siunitx}
\usepackage{enumitem}

\usepackage[normalem]{ulem} % for \uline

\usepackage{listings}
\usepackage[most]{tcolorbox}

\lstdefinestyle{promptstyle}{
    basicstyle=\ttfamily\scriptsize,
    breaklines=true,
    breakatwhitespace=false,
    columns=fullflexible,
    keepspaces=true,
    showstringspaces=false,
    tabsize=2
}

\newtcolorbox{promptbox}[1]{
    colback=gray!3,
    colframe=gray!45,
    title=#1,
    fonttitle=\bfseries\small,
    boxrule=0.4pt,
    arc=1.5pt,
    left=4pt,
    right=4pt,
    top=4pt,
    bottom=4pt,
    before skip=0.6em,
    after skip=0.6em
}

\newtcblisting{promptlisting}[1]{
    colback=gray!3,
    colframe=gray!45,
    title=#1,
    fonttitle=\bfseries\small,
    boxrule=0.4pt,
    arc=1.5pt,
    left=4pt,
    right=4pt,
    top=4pt,
    bottom=4pt,
    before skip=0.4em,
    after skip=0.4em,
    breakable,
    enhanced jigsaw,
    listing only,
    listing options={style=promptstyle}
}

\setlist[description]{%
  leftmargin=0pt,
  labelindent=0pt,
  labelsep=.6em,
  itemsep=.3\baselineskip
}

\begin{document}

\title{FitVTON: Fit-aware Virtual Try-On via Body-Garment Size Control}
\author{
  Yiqun Ning\textsuperscript{1}
  \quad Ao Shen\textsuperscript{1,2}
  \quad Chenhang He\textsuperscript{1}\thanks{Corresponding author.}
  \quad Lei Zhang\textsuperscript{1}\\
  \textsuperscript{1}Department of Computing, The Hong Kong Polytechnic University, Hong Kong\\
  \textsuperscript{2}Nuvatech\\
  \texttt{zeno-yq.ning@polyu.edu.hk, ashen@polyu.edu.hk, shenao@nuvatech.me}\\
  \texttt{chenhang.he@polyu.edu.hk, cslzhang@comp.polyu.edu.hk}
}

\maketitle

\begin{abstract}
    While diffusion-based virtual try-on has achieved impressive visual realism, most methods treat the task as 2D inpainting, prioritizing texture preservation over physical plausibility. Consequently, they often produce plausible-looking images that fail to reflect authentic garment fit across diverse body shapes. We present FitVTON, a Fit-aware virtual try-on model on different bodies in the wild. FitVTON encodes garment-body size through structured text prompts, and learn from simulated try-on triplets from parameterized garment model. To improve the fitting effects over garment silhouettes, we introduce two auxiliary heads to predict the masks for both the garment and the exposed body. We further introduce a texture rectification stage to improve realistic appearance from simulated data. To evaluate the fitting fidelity, we curate a real-world dataset, \textit{FittingEffect3K}, combining VLM-based scoring protocol. Both subjective and quantatitive experiments show that FitVTON demonstrates authentic fitting fidelity, with significant sizing accuracy and shape preservation over state-of-the-art methods while maintaining competitive image quality. Project Page: \url{https://zenoning.github.io/FitVTON/}.
\end{abstract}

\section{Introduction}
\label{sec:intro}
Virtual try-on (VTON) visualizes how a garment would look on a person, serving as a critical technology for e-commerce and content creation \cite{han2018viton,wang2018toward,yang2020towards,lewis2021tryongan}. Powered by diffusion models, modern VTON systems can synthesize compelling results with sharp textures and plausible lighting \cite{zhu2023tryondiffusion,choi2024improving,yang2024texture, li2023warpdiffusion}. Yet, visual realism does not guarantee physical authenticity: outputs may look convincing while failing to reflect how garments drape and fit across body shapes.

\begin{figure*}[h]
  \centering
  \includegraphics[width=\textwidth]{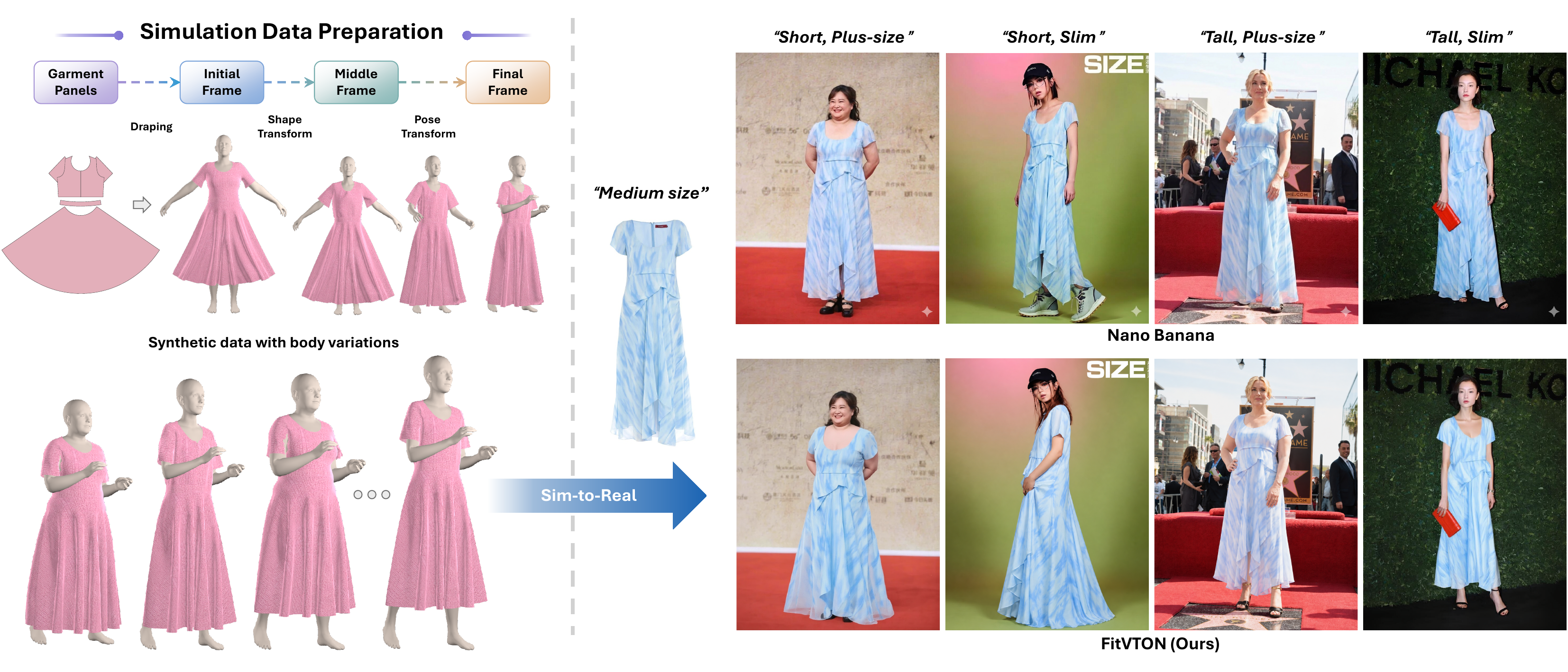}
  \caption{With garment-body size prompts, Nano Banana \cite{google2025nanobananapro} produce ``neutral fit" results across different body shape. In contrast, FitVTON produces more faithful fit results. (Zoom in to see how the hem position, tightness and the elasticity of the cuffs are varied across different bodies.)}
  \label{fig:one}
\end{figure*}

Figure~\ref{fig:one} illustrates this gap. Even when Nano Banana \cite{google2025nanobananapro} is given explicit garment and body size prompts, it produces "neutral fit" try-on results across different bodies. This suggests that fitting fidelity cannot be obtained from language prompts alone; models need supervision that captures how garments physically deform with body shape.

We identify two barriers to authentic fit. First, public datasets are skewed toward professional models with tall, slim figures \cite{choi2021viton,morelli2022dress,ge2019deepfashion2}, so models often generalize poorly to diverse bodies and default to an averaged fit that ignores individual characteristics. Second, current VTON datasets such as VITON-HD~\cite{choi2021viton} and DressCode~\cite{morelli2022dress} lack explicit \textit{(human, garment, target try-on)} triplets with controllable garment--body fit variation. As a result, most VTON approaches formulate the task as clothing-region inpainting \cite{choi2024improving,yang2024texture,kim2024stableviton,wan2025incorporating,li2024anyfit}. Although masks are effective for texture transfer and local realism, they can remove body-shape cues around fit-critical regions and let the prescribed mask boundary act as an implicit shape prior. The final result is therefore sensitive to mask quality and may follow the inpainting region rather than learning body-specific garment deformation.

We introduce \textbf{FitVTON}, a fit-aware virtual try-on model that generates more authentic fitting effects based on Garment-Body Size control. The model fine-tunes a \textit{FLUX.1 Kontext}~\cite{labs2025flux1kontextflowmatching} Image Editing model with simulated garment-fitting data, which encode cloth deformations across different body shapes and poses. As shown in Figure~\ref{fig:one}, by encoding the garment and body Size as structured text prompts (e.g., \textit{``medium-size skirt''} on \textit{``tall, plus-sized body''}), FitVTON produces try-on results that adapt more faithful garment cut, drape, and tightness under different body shapes and poses.

To generate diverse garment fitting data, we use GarmentCode \cite{korosteleva2023garmentcode} as our data source and SMPL-X \cite{pavlakos2019expressive, loper2023smpl} serves as our fitting mannequin. Both of them are parametric models, which allow us to control the garment-body size of the simulated data. We develop a multi-step simulation tools based on NVIDIA Warp \cite{macklin2022warp} that can generate garment fitting data with different body shapes, poses, and wearing styles, including one-piece garments, tucked-in/untucked upper outfits, and lower outfits.

To help the model learn fitting dynamics from simulated data, we introduce two innovations. First, we propose Fit-Grounded Dual-Branch Mask Supervision, which uses training-only auxiliary heads to predict both garment and exposed-body masks. With segmentation losses, these heads spatially ground the generated outputs to fit-sensitive silhouettes (e.g., expanded garment masks for an ``oversized'' cloth). Second, to bridge the sim-to-real domain gap, we introduce a texture rectification stage by fine-tuning the model on an unpaired real-world dataset. This enhances texture transferability while preserving the model’s fitting capability.

Finally, assessing the semantic and geometric accuracy of garment fit remains a significant challenge, as standard perceptual metrics (e.g., FID~\cite{heusel2017gans}, KID~\cite{binkowski2018demystifying}, LPIPS~\cite{zhang2018unreasonable}) prioritize global image fidelity over sizing correctness. To address this, we curate a specialized real-world benchmark, \textit{FittingEffect3K}, captured with 10 human participants with diverse body shapes, poses and cloth types, which produce ground-truth for direct evaluation of fitting accuracy. We design a multi-dimensional evaluation protocol that integrates both VLM-based scoring and subjective human preference. Empirically, we observe strong concordance between the VLM assessments and human judgments across various baselines. Both metrics consistently demonstrate that FitVTON significantly outperforms state-of-the-art methods, particularly in achieving accurate sizing and faithful shape preservation. In summary, our contributions are as follows:

\begin{itemize}
\item \textbf{Controllable Simulation Data Pipeline.} We generate large-scale, aligned try-on triplets with diverse body shapes, poses, and wearing styles.
\item \textbf{Fit-aware Flow Matching Model.} We fine-tune a FLUX-based model with dual-branch mask supervision and texture rectification for controllable and realistic fitting on various garment-body size..
\item \textbf{Fit Evaluation Dataset and Protocol.} We introduce the real-world \textit{FittingEffect3K} benchmark and evaluate fitting accuracy with VLM scoring and human assessment.
\end{itemize}

\section{Related Work}

\subsection{Image-based Virtual Try-On}
Image-based virtual try-on (VTON) aims to synthesize a person wearing a target garment while preserving both garment identity and human appearance. Early methods~\cite{han2018viton, wang2018toward, ge2021parser, xie2023gp} typically combine geometric warping with texture synthesis, and later warping-based pipelines improve consistency using appearance-flow constraints and local--global composition.
More recently, diffusion-based VTON methods~\cite{zhu2023tryondiffusion, kim2024stableviton, xu2025ootdiffusion, kim2025promptdresser, jiang2024fitdit,morelli2023ladi,zhang2024mmtryon,wan2025incorporating, yang2024d} achieve strong realism by formulating try-on as mask-guided inpainting conditioned on person, garment, and pose cues. However, cloth-agnostic masks may suppress body geometry near garment--body interfaces, weakening supervision on fit-critical boundaries (e.g., waistline, hips, sleeves, and hems). Recent mask-free VTON approaches increasingly build on large text-to-image backbones such as \textit{FLUX.1}~\cite{blackforestlabs2024flux}, leveraging its strong prompt understanding and image synthesis to remove explicit mask requirements at inference time~\cite{zhang2025boow, guo2025any2anytryon, feng2025omnitry, wang2025jco}. Despite improved usability and visual quality, these methods remain largely appearance-driven, and the language capability is not explicitly aligned with fit semantics or body--garment interactions. In contrast, FitVTON focuses on fitting fidelity rather than only appearance transfer. We fine-tune a FLUX-based model with simulation data that explicitly captures garment deformation under different body shapes and poses, enabling text prompts to control fit-related changes such as tightness, drape, and silhouette.

\subsection{Physics-Based Garment Simulation}
Physics-based cloth simulation provides a principled way to model garment deformation and body--garment interactions, and has started to be used in virtual fitting. Commercial systems~\cite{clo3d2022,style3d2022} achieve high fidelity but require expert-intensive garment authoring, limiting scalability. Recent work on procedural pattern representations, such as GarmentCode~\cite{korosteleva2023garmentcode,korosteleva2024garmentcodedata}, significantly reduces garment creation cost and enables large-scale synthesis, with further extensions enriching garment diversity and semantic control~\cite{bian2025chatgarment,zhou2025design2garmentcode}. Nonetheless, simulation pipelines are typically not optimized for photorealistic rendering and struggle to cover the appearance diversity needed for image-based VTON. Building on these efforts, we extend GarmentCode-based dynamic simulation to generate controllable, fitting-aware supervision across body shapes, and use it to train an image-based try-on model. This design combines the physical realism of simulation with the scalability and visual fidelity of generative image models, yielding photorealistic results with explicit fitting behavior.

\section{Method}
\label{sec:method}
\subsection{Garment-Body Size and Simulation}

Image-based VTON cannot reliably observe absolute garment sizes like S/M/L/XL; the labeling standards are often uncertain across brands, garment categories, and styling preferences. Instead of modeling absolute garment size, we describe fit as a relative relation between the body and the try-on garment. We refer to this relative relation as \textit{Garment-Body Size}. This formulation avoids relying on ambiguous garment-size annotations while directly capturing the visual fit changes that matter for try-on. To instantiate Garment-Body Size, we construct 16 representative SMPL-X body shapes covering different height and body-size combinations such as \textit{(tall, slim)}, \textit{(medium-tall, plus-size)}, etc. For each garment, GarmentCode \cite{korosteleva2023garmentcode} provides a medium-size pattern fitted to the mean body of SMPL-X \cite{pavlakos2019expressive,loper2023smpl}. For arbitrary garment patterns and body shapes, the same representation can be obtained by comparing body measurements with the corresponding measurement lines on the garment pattern, such as chest, waist, hip, garment length, and sleeve length, and use them as metrics to interpolate over our 16 predefined Garment-Body Size prototypes.

We use NVIDIA Warp \cite{macklin2022warp} with XPBD-based cloth simulation to simulate the garment on the body. The body first morphs from the mean shape to the target shape and then articulates to the target pose; we subdivide these transitions when needed to reduce cloth--body penetration. The resulting simulations provide controlled relative fit variations under a shared garment pattern, allowing the model to learn how natural-language prompts such as \textit{slim}, \textit{medium-tall}, and \textit{plus-sized} correspond to garment-body fit. We additionally model the outfit styles within three cases: one-piece, tucked-in, and untucked. For tucked-in/untucked outfits, the simulation uses separate upper and lower garment meshes and controls whether the upper hem lies inside/outside the lower waistband. Besides, we add wrist/ankle stop constraints to prevent sleeve cuffs or pant hems from sliding beyond the hand or shoe region to align with the real-world try-on scenario. The garment-body size prototypes and simulation details are elaborated in the appendix.

\subsection{FitVTON Architecture}

\begin{figure*}[t]
  \centering
  \includegraphics[width=\linewidth]{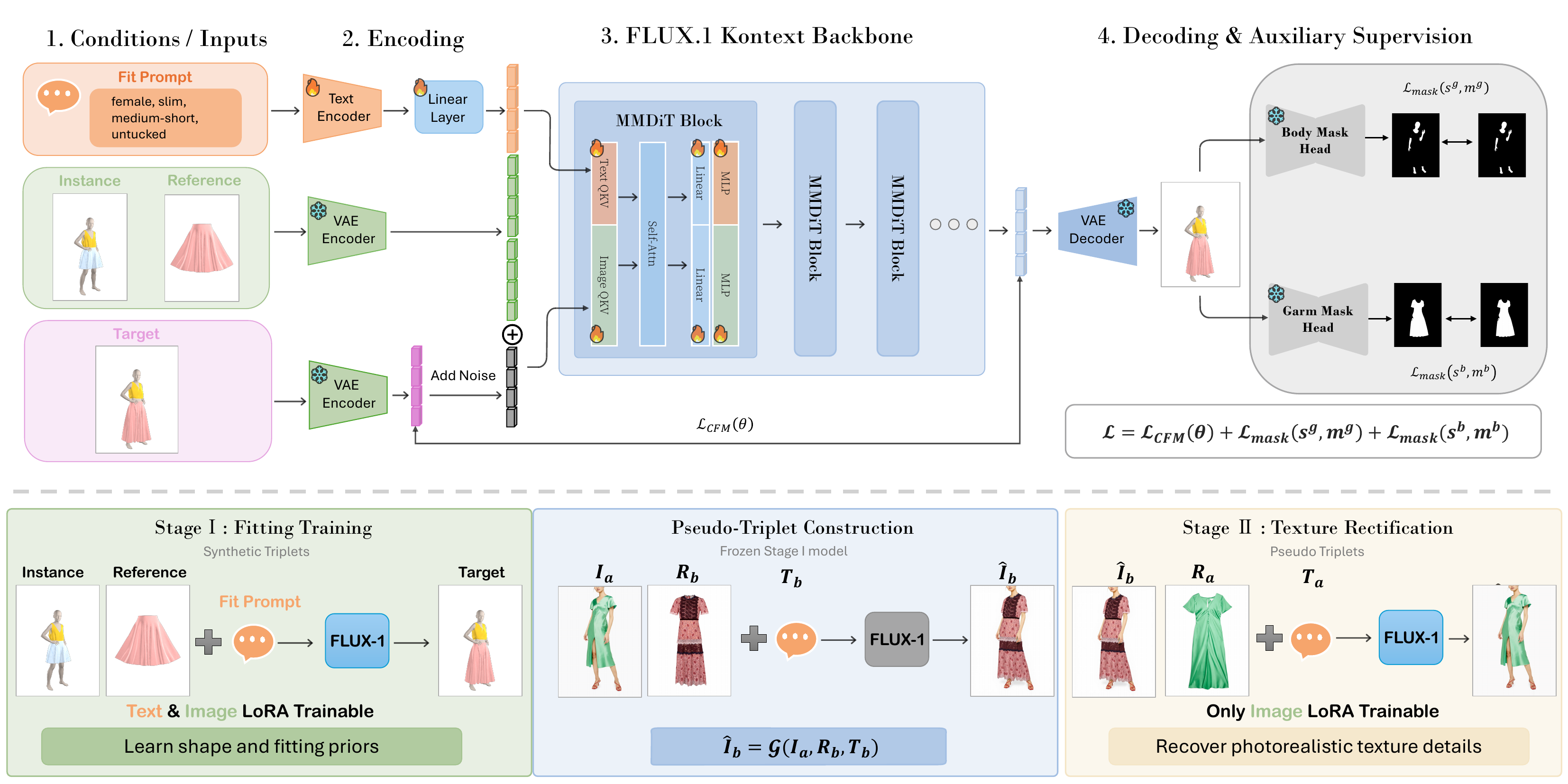}
   \caption{Overview of FitVTON. (\textbf{Top}) Given a person image, a reference garment, and a Garment-Body Size prompt, FitVTON performs fit-aware virtual try-on with FLUX.1 Kontext backbone, modality-specific LoRA adapters, and dual-branch garment/body mask supervision. (\textbf{Bottom}) Two-stage training strategy: Stage I learns prompt-driven fit priors from synthetic simulations, while Stage II rectifies real-image textures by updating only image LoRA layers.}
  \label{fig:three}
\end{figure*}

As shown in Fig.~\ref{fig:three}, FitVTON is formulated as fit-aware image editing. The input conditions consist of an instance image $I_a$, a reference garment image $R_b$, and a fit prompt $\mathbf{T}_b$ that describes Garment-Body Size and wearing style, e.g., \textit{female, slim, medium-short, untucked}. The model generates a target try-on image $I_b$ that preserves the visual appearance of $R_b$ while adapting the garment silhouette to the body and pose in $I_a$ according to the fit prompt.

We build FitVTON on \textit{FLUX.1 Kontext}~\cite{labs2025flux1kontextflowmatching}, a flow-matching Multimodal Diffusion Transformer. The instance image, reference garment, and target image are encoded into the latent space by the VAE encoder, while the prompt is encoded by the text encoder. These visual and textual tokens are then processed by the MMDiT backbone, where image and text LoRA adapters are inserted into the modality-specific projections. This design lets the image branch learn garment transfer and appearance preservation, while the text branch learns how Garment-Body Size controls fit-related geometry. Let $\mathcal{E}$ and $\mathcal{D}$ denote the VAE encoder and decoder. For a training triplet $(I_a, R_b, I_b)$, we encode the target try-on image as $\mathbf{x}_1=\mathcal{E}(I_b)$ and sample a noise latent $\mathbf{x}_0\sim p_0$. The conditioning signal is denoted by $\mathbf{c}$ and contains the encoded instance image, reference garment, and fit prompt:
\begin{equation}
    \mathbf{c} = \left\{\mathcal{E}(I_a), \mathcal{E}(R_b), \tau(\mathbf{T}_b)\right\},
\end{equation}
where $\tau(\cdot)$ is the text encoder. The flow-matching backbone learns a conditional vector field governed by the ODE:
\begin{equation}
\frac{d\mathbf{x}_t}{dt} = \mathbf{v}_\theta(\mathbf{x}_t, t, \mathbf{c}), \quad t\in[0,1],
\end{equation}
where $\mathbf{x}_t$ is the intermediate latent state and $\mathbf{v}_\theta$ predicts its instantaneous velocity. We use the linear probability path between the noise and data endpoints:
\begin{equation}
\mathbf{x}_t = (1-t)\mathbf{x}_0 + t\mathbf{x}_1,
\end{equation}
whose target velocity is $\mathbf{u}_t=\mathbf{x}_1-\mathbf{x}_0$. The conditional flow-matching objective is:
\begin{equation}
\mathcal{L}_{\text{CFM}}(\theta) =
\mathbb{E}_{t\sim\mathcal{U}(0,1),\, \mathbf{x}_0\sim p_0,\, \mathbf{x}_1\sim p_{\text{data}}}
\left[
\left\lVert \mathbf{v}_\theta(\mathbf{x}_t, t, \mathbf{c}) - \mathbf{u}_t \right\rVert_2^2
\right].
\end{equation}
At inference, the learned ODE is integrated from a sampled noise latent to obtain $\hat{\mathbf{x}}_1$, which is decoded as $\hat{I}_b=\mathcal{D}(\hat{\mathbf{x}}_1)$. While this objective provides the main image-generation supervision, it does not explicitly constrain whether the fit prompt controls local garment and body boundaries. We address this limitation with the auxiliary supervision below.

\subsection{Dual-Branch Mask Supervision}
Fit variations are expressed most clearly through local geometry, including waistlines, hems, sleeves, cuffs, garment boundaries, and exposed-body regions. A pure flow-matching objective can reproduce the target appearance, but it provides only indirect supervision for whether each intermediate state follows the intended fit semantics. We therefore introduce Dual-Branch Mask Supervision as an auxiliary geometric regularizer. From the simulated training data, we export garment masks $\mathbf{M}^{g}$ and exposed-body masks $\mathbf{M}^{b}$, and use them to train two lightweight U-Net mask heads, $\mathcal{H}_{\phi}^{g}$ and $\mathcal{H}_{\psi}^{b}$. During flow-matching training, the predicted velocity at timestep $t$ gives a one-step estimate of the endpoint latent:
\begin{equation}
    \tilde{\mathbf{x}}_{1}^{(t)}
    = \mathbf{x}_t + (1-t)\mathbf{v}_\theta(\mathbf{x}_t, t, \mathbf{c}).
\end{equation} 
We decode $\tilde{\mathbf{x}}_{1}^{(t)}$ to image space and apply the two mask heads to predict timestep-conditioned geometry:
\begin{equation}
    \hat{\mathbf{M}}_{t}^{g}
    = \mathcal{H}_{\phi}^{g}\!\left(\mathcal{D}(\tilde{\mathbf{x}}_{1}^{(t)})\right),
    \quad
    \hat{\mathbf{M}}_{t}^{b}
    = \mathcal{H}_{\psi}^{b}\!\left(\mathcal{D}(\tilde{\mathbf{x}}_{1}^{(t)})\right),
\end{equation}
where $\mathcal{D}$ denotes the image decoder. Each branch is supervised by a BCE--Dice objective:
\begin{equation}
    \mathcal{L}_{\text{seg}}(\hat{\mathbf{M}}, \mathbf{M}) =
    \mathcal{L}_{\text{BCE}}(\hat{\mathbf{M}}, \mathbf{M}) +
    \eta\mathcal{L}_{\text{Dice}}(\hat{\mathbf{M}}, \mathbf{M}),
\end{equation}
where $\eta$ balances binary cross-entropy and Dice loss \cite{sudre2017generalised}. The final training objective is:
\begin{equation}
    \mathcal{L}_{\text{total}} =
    \mathcal{L}_{\text{CFM}}
    + \lambda_g \mathcal{L}_{\text{seg}}(\hat{\mathbf{M}}_{t}^{g}, \mathbf{M}^{g})
    + \lambda_b \mathcal{L}_{\text{seg}}(\hat{\mathbf{M}}_{t}^{b}, \mathbf{M}^{b}).
\end{equation}
The garment branch encourages prompt-consistent clothing silhouettes, while the exposed-body branch preserves body-shape cues that should remain visible under different fit prompts. Since the U-Net heads are used only as training-time regularizers and are removed at inference, FitVTON does not require masks at test time.

\subsection{Texture Rectification}
As illustrated in Fig.~\ref{fig:three}, Stage I learns the mapping from fit prompts to garment-body geometry on physically simulated data. This stage provides reliable fit supervision, but synthetic renderings still differ from real fashion images in texture complexity, lighting, fabric patterns, and high-frequency details. Directly fine-tuning the whole model on real try-on data may improve appearance, but it can also weaken the Garment-Body Size semantics learned from simulation. We therefore introduce a second-stage Texture Rectification procedure that adapts the model to real-image appearance while preserving the prompt-driven fit control from Stage I.

Texture Rectification follows the modality-specific LoRA design shown in Fig.~\ref{fig:three}. We freeze the text-modality LoRA layers, which encode the fit-prompt prior, and update only the image-modality LoRA layers responsible for visual conditioning and texture transfer. This restricts the second stage to correcting real-image appearance instead of relearning the text-to-fit mapping.

We use VITON-HD~\cite{choi2021viton} and DressCode~\cite{morelli2022dress} for rectification. For an unpaired real sample, the fixed stage-one model $\mathcal{G}_{\text{stage1}}$ generates a pseudo target from a person image $I_a$, a reference garment $R_b$, and the corresponding prompt $\mathbf{t}_b$:
\begin{equation}
\hat{I}_b = \mathcal{G}_{\text{stage1}}(I_a, R_b, \mathbf{t}_b),
\end{equation}
forming a pseudo-triplet $(I_a, R_b, \hat{I}_b)$. These pseudo-triplets preserve the garment-exchange task structure because the input still contains a person image and a separate reference garment. However, pseudo targets alone may inherit artifacts from the stage-one model. We therefore mix them with real reconstruction pairs $(I_a, R_a, I_a)$, where the model is asked to reconstruct the original person image from its paired garment. The reconstruction pairs anchor the training to real textures, identity details, and high-frequency garment appearance, while the pseudo-triplets maintain the try-on transfer behavior. Since the text-modality layers remain frozen throughout this stage, Texture Rectification improves realism without overwriting the fit semantics established by the simulation stage.

\section{Experiment}
We leave all the implementation details of training and inference in Appendix.

\subsection{Datasets and Evaluation Setup}

\noindent \textbf{Dataset.} Stage I LoRA training uses GarmentCodeVTON, a synthetic triplet dataset generated by our dynamic simulation pipeline, which combines 19 garment references, 16 female and male body-shape variations, and 10 pose variations, producing 78{,}080 training triplets. Stage II LoRA finetuning is conducted on pseudo triplets, including 1{,}000 triplets from VITON-HD~\cite{choi2021viton} and 3{,}000 triplets from DressCode~\cite{morelli2022dress}.

For evaluation, we use public VITON-HD and DressCode unpair test splits for image-quality assessment, and introduce \textit{FittingEffect3K} for fit-oriented evaluation. It contains 3{,}350 real-world evaluation triplets formed from 14 medium-sized garments containing corresponding measurements, 5 male models and 5 female models in 5 different poses, together with their body measurements. 

\begin{figure*}[t]
    \centering
    \begin{minipage}[t]{0.47\textwidth}
        \centering
        \includegraphics[width=\linewidth]{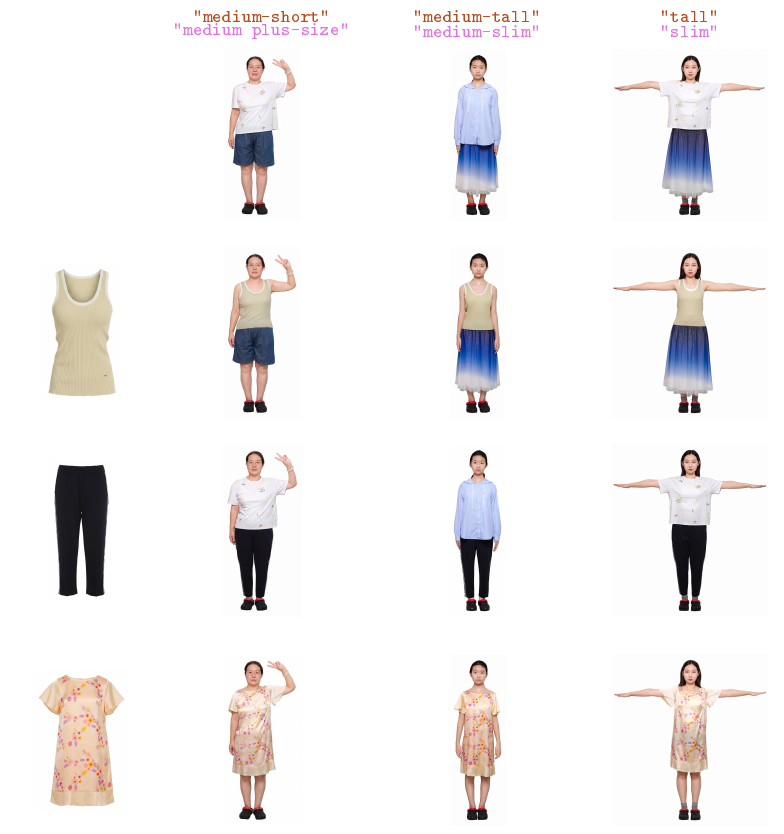}
    \end{minipage}\hfill
    \begin{minipage}[t]{0.53\textwidth}
        \centering
        \includegraphics[width=\linewidth]{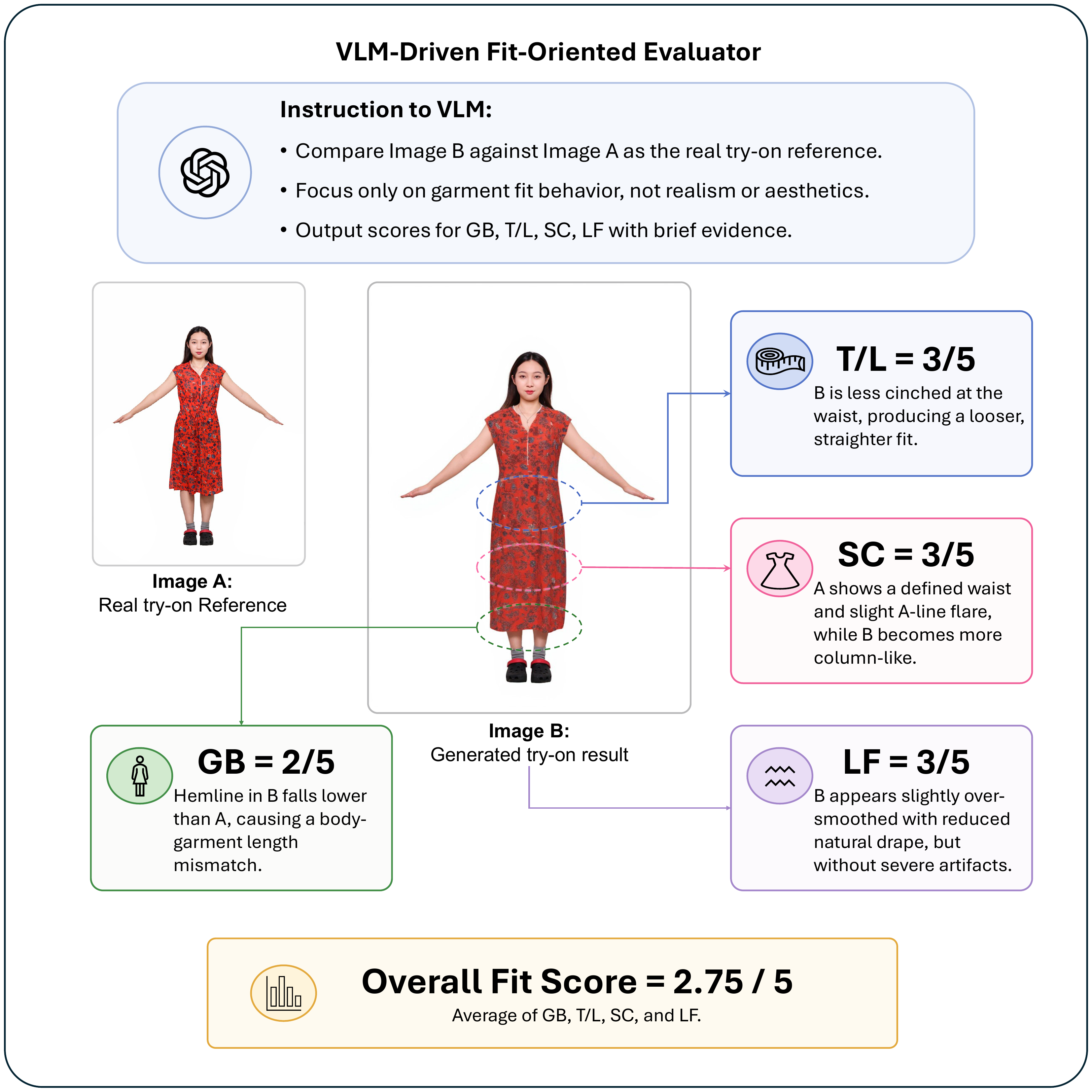}
    \end{minipage}
    \caption{\textbf{Left:} Samples of the \textit{FittingEffect3K} dataset, a real-world benchmark for evaluating fit-aware virtual try-on across diverse body shapes, poses, and garments. \textbf{Right:} Our fit-oriented VLM evaluation protocol. Given a real try-on reference and a generated result, the evaluator scores four fit dimensions (GB, T/L, SC, LF; 1--5) with brief evidence and reports the averaged fit score.}
    \label{fig:FittingEffect3K}
\end{figure*}

\begin{figure*}[t]
  \centering
  \includegraphics[width=\textwidth]{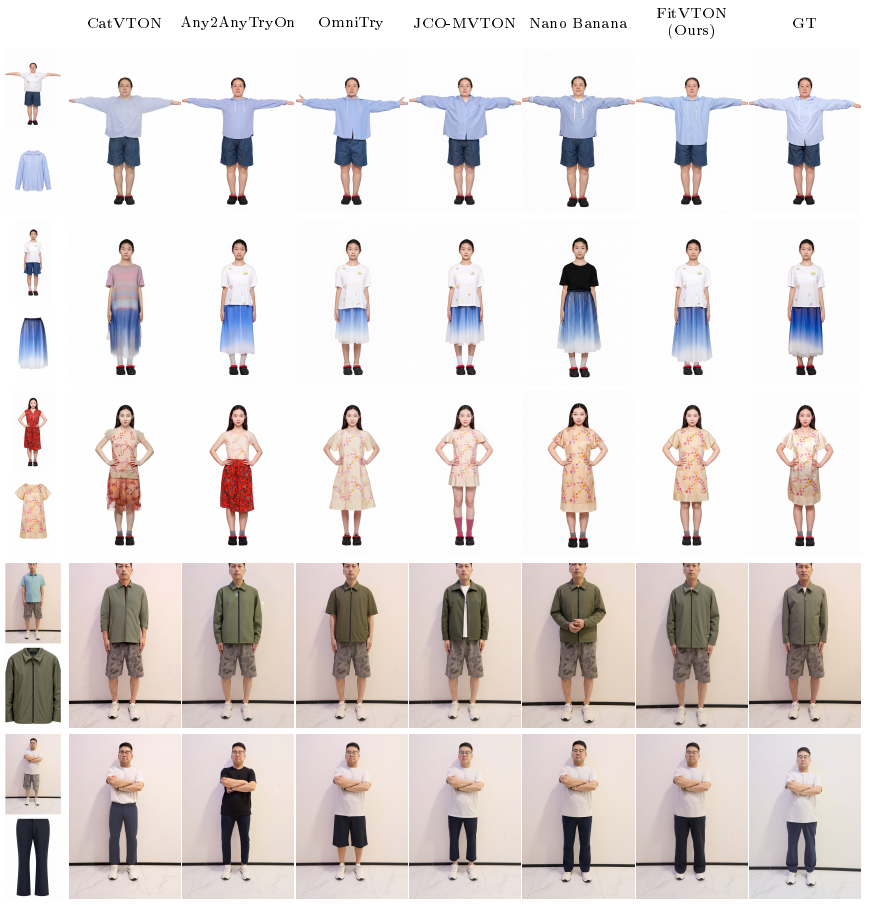}
  \vspace{-5mm}
  \caption{Qualitative results on \textit{FittingEffect3K}.}
  \label{fig:qualresultFittingEffect3K}
\end{figure*}

\noindent \textbf{Fit-Oriented Evaluation Protocol.} 
We evaluate garment fitting quality using a fit-oriented protocol, as existing metrics correlated to virtual try-on primarily target visual fidelity or semantic similarity and are not designed to reliably assess fitting behavior. Our protocol decomposes fitting behavior into four complementary dimensions: (i) Garment--Body Alignment (GB), which measures the consistency of garment placement relative to anatomical landmarks; (ii) Tightness/Looseness Consistency (T/L), which assesses whether the relative ease of the garment around different body regions matches the real try-on reference; (iii) Silhouette Consistency (SC), which captures the preservation of the overall garment-induced body silhouette; and (iv) Local Fit Artifacts (LF), which identify localized fitting failures such as unnatural pulling, wrinkling, fabric collapse, or deformation. We use GPT-5.2 as a scoring agent for scalable assessment. Given a ground-truth try-on image and a generated result of the same garment on the same body shape, the evaluator treats the former as the reference and rates the latter from 1 to 5 on each fitting dimension, focusing exclusively on fit accuracy rather than realism or aesthetics. We report per-dimension, category-wise, and overall average scores, using category-specific prompts for upper-body, lower-body, and dress try-on. A representative scoring example is shown in Figure~\ref{fig:FittingEffect3K} (Right). More prompt details and the calibration analysis are provided in the appendix.

\noindent \textbf{Other Metrics.}
We further report subjective human evaluation and general image-quality metrics on \textit{FittingEffect3K}. For subjective evaluation, we conduct a human preference study in which the ground-truth reference and generated results from 6 methods are randomly shuffled, and each participant selects the result that best matches the reference in terms of garment fit. Specifically, we recruit 20 participants to evaluate 100 comparison cases, resulting in 2000 selections. For image quality, we use FID~\cite{heusel2017gans} and KID~\cite{binkowski2018demystifying} to measure distribution-level similarity between generated results and ground-truth images.

\begin{table*}[t]
  \centering
  \caption{\textbf{Fit-Oriented Evaluation Protocol.}
  We report four fit dimensions: GB (Garment--Body Alignment), T/L (Tightness/Looseness),
  SC (Silhouette Consistency), and LF (Local Fit Artifacts).}
  \label{tab:fit_protocol_full}

  \scriptsize
  \setlength{\tabcolsep}{1.2pt}
  \renewcommand{\arraystretch}{1.2}

  \resizebox*{\textwidth}{!}{%
    \begin{tabular}{@{}>{\raggedright\arraybackslash}m{2.8cm} ccccc ccccc ccccc c@{}}
      \toprule
      \multirow{2}{*}{Method} &
      \multicolumn{5}{c}{Upper} &
      \multicolumn{5}{c}{Lower} &
      \multicolumn{5}{c}{Dress} &
      \multirow{2}{*}{Whole} \\
      \cmidrule(lr){2-6} \cmidrule(lr){7-11} \cmidrule(lr){12-16}
      & GB & T/L & SC & LF & Avg
      & GB & T/L & SC & LF & Avg
      & GB & T/L & SC & LF & Avg
      & Avg \\

      \midrule

      CatVTON~\cite{chong2024catvton}
      & 2.78 & 2.55 & 2.42 & 2.74 & 2.62
      & 2.10 & 1.95 & 1.85 & 2.46 & 2.09
      & 1.90 & 2.00 & 1.75 & 2.15 & 1.95
      & 2.30 \\

     Any2AnyTryOn~\cite{guo2025any2anytryon}
      & 3.03 & 2.70 & 2.65 & 3.32 & 2.92
      & 2.47 & 2.21 & 2.23 & 2.98 & 2.47
      & 1.50 & 1.75 & 1.45 & 2.45 & 1.79
      & 2.57 \\

      OmniTry~\cite{feng2025omnitry}
      & 3.20 & 2.78 & 2.77 & 3.26 & 3.00
      & 1.93 & 2.04 & 1.79 & 2.85 & 2.15
      & 2.50 & 2.25 & 2.25 & 2.60 & 2.40
      & 2.55 \\

      JCo-MVTON~\cite{wang2025jco}
      & 3.13 & 2.71 & 2.64 & 3.38 & 2.96
      & 2.75 & 2.41 & 2.41 & 3.26 & 2.71
      & 1.80 & 2.25 & 1.95 & 2.60 & 2.15
      & 2.74 \\

      \midrule
  
      Nano Banana\cite{google2025nanobananapro}
      & 3.23 & 2.91 & 2.86 & \textbf{3.75} & 3.19
      & 2.33 & 2.13 & 2.34 & 2.98 & 2.45
      & 2.90 & 2.60 & 2.45 & \textbf{3.35} & 2.83
      & 2.82 \\
      
      \midrule

      FitVTON
      & \textbf{3.40} & \textbf{2.94} & \textbf{2.87} & 3.69 & \textbf{3.22}
      & \textbf{3.20} & \textbf{2.68} & \textbf{2.72} & \textbf{3.37} & \textbf{2.99}
      & \textbf{2.95} & \textbf{2.65} & \textbf{2.80} & 3.20 & \textbf{2.90}
      & \textbf{3.08} \\

FitVTON (w/o mask)
& 3.06 & 2.71 & 2.68 & 3.39 & 2.96
& 2.83 & 2.52 & 2.40 & 3.23 & 2.79
& 2.90 & 2.55 & 2.65 & 3.15 & 2.81
& 2.87 \\

      \bottomrule
    \end{tabular}%
  }
\end{table*}

\subsection{Comparison with Baselines}
We compare FitVTON with recent mask-free virtual try-on test baselines through qualitative and quantitative evaluations, including several state-of-the-art open-source methods~\cite{chong2024catvton,guo2025any2anytryon,feng2025omnitry,wang2025jco} and the commercial image editing model \cite{google2025nanobananapro}. To ensure a fair comparison and allow each baseline to produce its best try-on results, we use the default prompts provided by the respective methods. We focus the main paper on fitting fidelity, and provide additional comparisons of public benchmark image quality. More visualizations are provided in the appendix.

To assess garment fitting correctness beyond appearance-level realism, we evaluate all methods on \textit{FittingEffect3K} using our proposed fit-oriented protocol, reporting category-wise scores for upper-body, lower-body, and dress try-on, together with category averages and an overall score. As shown in Table~\ref{tab:fit_protocol_full}, FitVTON achieves the best overall fitting consistency with the highest whole average score of 3.08. It consistently outperforms all academic baselines across garment categories, achieving average scores of 3.22 on Upper, 2.99 on Lower, and 2.90 on Dress. These improvements indicate better garment--body alignment and silhouette preservation relative to anatomical landmarks, leading to more reliable fitting behavior. Notably, even compared with the commercial model Nano Banana, FitVTON still obtains a higher overall fit score (3.08 vs. 2.82). We note that Nano Banana obtains particularly high Local Fit Artifact (LF) scores in some categories. This is because LF mainly penalizes visible local distortions such as tearing, collapse, or unnatural wrinkles; a commercial model can produce smooth and visually plausible garments with few such artifacts.

We further validate this trend through the human preference study in Table~\ref{tab:human_preference}, where FitVTON is selected most frequently among all compared methods. This suggests that the proposed fit-oriented evaluation protocol is broadly consistent with human perception of garment fitting quality under real try-on references. Qualitative comparisons are shown in Figure~\ref{fig:qualresultFittingEffect3K}.

\begin{figure*}[t]
  \centering
  \begin{minipage}[t]{0.5\textwidth}
    \centering
    \includegraphics[width=\linewidth]{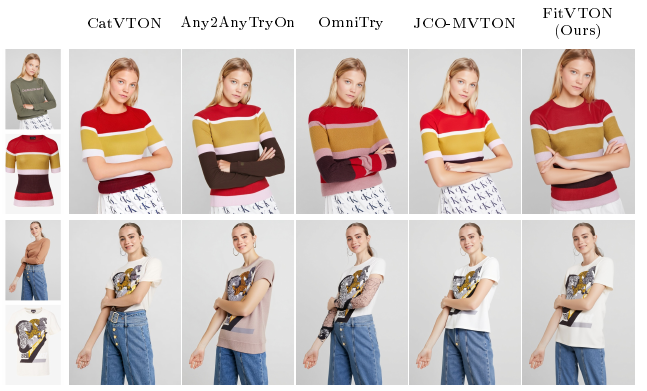}
  \end{minipage}\hfill
  \begin{minipage}[t]{0.5\textwidth}
    \centering
    \includegraphics[width=\linewidth]{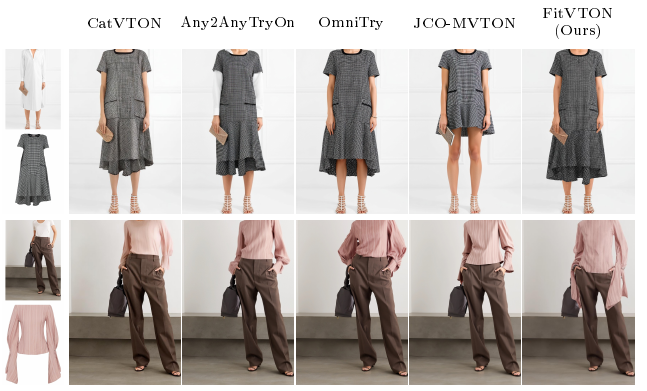}
  \end{minipage}
  \caption{Qualitative comparison on public benchmarks. Left: VITON-HD. Right: DressCode. }
  \label{fig:supp_qualresults}
\end{figure*}
Following the commonly adopted unpaired evaluation setting on VITON-HD and DressCode, we measure distribution-level similarity between generated and ground-truth images using FID~\cite{heusel2017gans} and KID~\cite{binkowski2018demystifying}. As shown in Table~\ref{tab:supp_fid_kid_comparison}, FitVTON achieves competitive FID/KID scores on DressCode (FID=5.21, KID=1.21), indicating that the proposed fit-aware design maintains general image realism. The comparison in Figure~\ref{fig:supp_qualresults} further shows that FitVTON preserves realistic appearance on standard public VTON benchmarks while targeting the complementary goal of fitting fidelity.

Meanwhile, FitVTON does not always achieve the best FID/KID among all baselines, which is expected given our optimization target. FID/KID measure distribution-level similarity, therefore tend to favor methods that closely match the overall appearance statistics of public VTON benchmarks. FitVTON explicitly adjusts garment silhouettes, tightness, and body--garment correspondence according to the fit prompt, some fit-correct geometric changes may not be rewarded by FID/KID.

\begin{figure*}[t]
  \centering
  \makebox[\textwidth][c]{%
    \begin{minipage}[t]{0.355\textwidth}
      \centering
      \includegraphics[width=\linewidth]{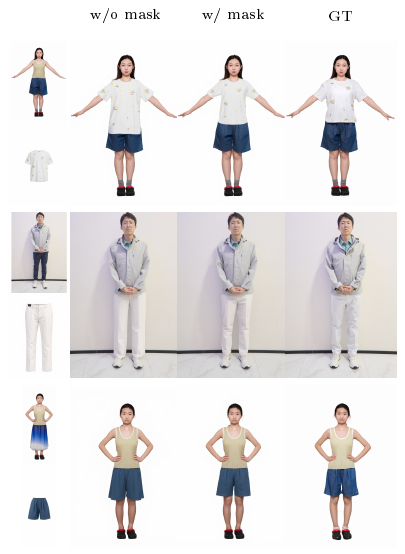}
    \end{minipage}%
    \hspace{0.04\textwidth}%
    \begin{minipage}[t]{0.26\textwidth}
      \centering
      \includegraphics[width=\linewidth]{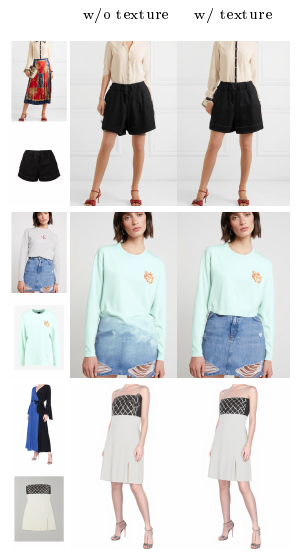}
    \end{minipage}%
  }
  \vspace{-5mm}
  \caption{Ablation results of the dual-branch mask supervision (left) and texture rectification (right). }
  \label{fig:Ablation}
\end{figure*}

\begin{table*}[t]
  \centering
  \setlength{\tabcolsep}{5pt}
  \renewcommand{\arraystretch}{1.12}

  \begin{minipage}[t]{0.42\textwidth}
    \centering
    \caption{Human preference study.}
    \label{tab:human_preference}

    \resizebox{\linewidth}{!}{%
    \begin{tabular}{lcc}
      \toprule
      Method & Selections$\uparrow$ & Ratio$\uparrow$ \\
      \midrule
      Nano Banana  & 517 & 25.85\% \\
      CatVTON      & 86  & 4.30\%  \\
      Any2AnyTryOn & 147 & 7.35\%  \\
      OmniTry      & 163 & 8.15\%  \\
      JCo-MVTON    & 421 & 21.05\% \\
      FitVTON      & \textbf{666} & \textbf{33.30\%} \\
      \bottomrule
    \end{tabular}%
    }
  \end{minipage}
  \hfill
  \begin{minipage}[t]{0.54\textwidth}
    \centering
    \caption{Image quality assessment. }
    \label{tab:supp_fid_kid_comparison}

    \resizebox{\linewidth}{!}{%
    \begin{tabular}{lcccc}
      \toprule
      Method & \multicolumn{2}{c}{VITON-HD} & \multicolumn{2}{c}{DressCode} \\
      \cmidrule(lr){2-3}\cmidrule(lr){4-5}
      & FID$\downarrow$ & KID$\downarrow$ & FID$\downarrow$ & KID$\downarrow$ \\
      \midrule
      CatVTON      
      & 9.9667 & 1.8646  
      & 6.2598 & 1.5676 \\
      Any2AnyTryOn 
      & \textbf{9.1742} & \textbf{1.1850}  
      & \textbf{4.8287} & \textbf{0.7579} \\
      OmniTry      
      & 11.6031 & 2.3572 
      & 5.3977 & 1.0283 \\
      JCo-MVTON    
      & 12.4363 & 3.9295 
      & 8.8442 & 3.5624 \\
      FitVTON      
      & 13.6398 & 4.8988   
      & 5.2105 & 1.2078 \\
      FitVTON (w/o texture)  
      & 16.1519 & 7.0423  
      & 11.6624 & 7.7163 \\
      \bottomrule
    \end{tabular}%
    }
  \end{minipage}
\end{table*}

\subsection{Ablation Study}

\begin{description}

\item[\textbf{Effect of the Dual-Branch Mask Supervision.}]
As shown in Figure~\ref{fig:Ablation} (Left), we investigate the impact of the proposed dual-branch mask module in the first-stage training on our GarmentCodeVTONDataset, and evaluate its fitting behavior on \textit{FittingEffect3K} using the fit-oriented protocol (Table~\ref{tab:fit_protocol_full}). Compared to the variant without this module, FitVTON improves overall fitting consistency, increasing the whole average score from 2.87 to 3.08. The improvement is consistent across garment categories, with notable gains on Upper (from 2.96 to 3.22), Lower (from 2.79 to 2.99) and Dress (from 2.81 to 2.90). These results indicate that the mask-guided dual-branch design helps the model better localize garment regions and better maintain alignment cues, improving fitting consistency in virtual try-on. More visualizations are provided in the appendix.

\item[\textbf{Effect of the Texture-Rectification.}]
We assess the contribution of the second-stage texture-rectification module, which adapts the simulation-trained model to real fashion images using VITON-HD and DressCode. As illustrated in Figure~\ref{fig:Ablation} (Right), removing this module substantially weakens image realism and distribution-level similarity. On VITON-HD, the FID/KID scores worsen from 13.64/4.90 to 16.15/7.04; on DressCode, they worsen from 5.21/1.21 to 11.66/7.72. These results show that texture rectification is effective in recovering high-frequency garment details, fabric patterns, and real-image appearance statistics that are not fully captured by the first-stage fitting-aware generation alone. More visualizations are provided in the appendix.

\end{description}

\section{Conclusion}

We have presented \textbf{FitVTON}, a fit-aware virtual try-on framework that addresses the gap between photorealistic synthesis and authentic garment fitting. FitVTON learns prompt-driven Garment-Body Size control from controllable simulation triplets, grounds fit-sensitive geometry with dual-branch mask supervision, and uses texture rectification to improve realism on real fashion images. Together with \textit{FittingEffect3K} and a fit-oriented evaluation protocol, our experiments show that FitVTON improves sizing accuracy and body-shape preservation over strong mask-free and commercial baselines while maintaining competitive image quality. These results highlight physically grounded fit supervision as a promising direction for practical virtual try-on.

\noindent \textbf{Limitations.} FitVTON controls fitting effects through Garment-Body Size prompts derived from 16 representative body-size prototypes. While this captures broad fit variations across body categories, the control granularity is still coarse and cannot specify continuous body measurements, garment dimensions, or centimeter-level ease. Finer-grained quantitative size control remains future work.

\clearpage  % TODO FINAL: This \clearpage needs to be removed from both review and camera-ready versions.

% ---- Bibliography ----
%
% NeurIPS template commonly uses plainnat.
\bibliographystyle{plainnat}
\bibliography{main}

\clearpage

\input{supp}

\end{document}

%% file: supp.tex
% Appendix content input by main.tex. Keep this file free of documentclass,
% package imports, title, author, bibliography, and end{document}.
\appendix

\begin{center}
    \Large\textbf{Appendix}\par
\end{center}

In this appendix, we provide supplementary materials for FitVTON. 
\begin{enumerate}
\item[\textbf{A.}] Details of the simulation pipeline corresponding to Sec.~3.1 of the main paper, including body-motion scheduling, dynamic cloth updates, wearing-style control, and limb-end constraints.
\item[\textbf{B.}] Details of the implementation settings corresponding to Sec.~4 of the main paper, including training, LoRA adaptation, and inference.
\item[\textbf{C.}] Details of the VLM-based fitting evaluator corresponding to Sec.~4.1 of the main paper, including repeated-run consistency analysis, explanation consistency analysis, and prompt design.
\item[\textbf{D.}] Visualization results supporting Secs.~4.2 and 4.3 of the main paper, including prompt-based fitting analysis, additional baseline comparisons, and ablation visualizations.
\end{enumerate}

\section{Simulation Pipeline and Garment-Body Prototypes}
\label{sec:supp_simulation}

\subsection{Two-Stage Garment Body Simulation}
\label{sec:supp_simulation_schedule}

Our pipeline combines parametric garment generation with physics-based dynamic cloth simulation. We use GarmentCode~\cite{korosteleva2023garmentcode} to generate 2D sewing patterns and garment structures, and NVIDIA Warp~\cite{macklin2022warp} to run XPBD-based cloth simulation. Each garment is first stitched and simulated on a default body with average shape in a canonical A-pose until it reaches static equilibrium, producing a stable initial draping state before any target-specific body motion is applied. We then split the body transition into shape morphing followed by pose articulation. Let $\boldsymbol{\theta}_s$ and $\boldsymbol{\beta}_s$ denote the start pose and shape, and let $\boldsymbol{\theta}_e$ and $\boldsymbol{\beta}_e$ denote the target pose and shape. The start state is the default A-pose body, and the end state is the target body shape and pose. We first interpolate $\boldsymbol{\beta}_s \rightarrow \boldsymbol{\beta}_e$ while keeping $\boldsymbol{\theta}_s$ fixed, and then interpolate $\boldsymbol{\theta}_s \rightarrow \boldsymbol{\theta}_e$ while keeping $\boldsymbol{\beta}_e$ fixed.

\begin{center}
\begin{minipage}{0.75\linewidth}
\begin{algorithm}[H]
\caption{Building the dynamic body sequence}
\label{alg:smooth_body_sequence}
\DontPrintSemicolon
\SetKwInOut{KwIn}{Input}
\SetKwInOut{KwOut}{Output}
\KwIn{SMPL-X model $M$; start pose/shape $(\boldsymbol{\theta}_s,\boldsymbol{\beta}_s)$;
target pose/shape $(\boldsymbol{\theta}_e,\boldsymbol{\beta}_e)$; refinement threshold $\tau$.}
\KwOut{A smooth body-vertex sequence $\mathcal{B}=\{\mathbf{B}_t\}_{t=0}^{T-1}$ for dynamic cloth simulation.}

\BlankLine
\tcp{Stage 1: change body shape while keeping the A-pose fixed}
$\mathcal{B}_{\beta} \leftarrow \textsc{AdaptiveInterpolate}(M,(\boldsymbol{\theta}_s,\boldsymbol{\beta}_s),(\boldsymbol{\theta}_s,\boldsymbol{\beta}_e),\tau)$\;

\BlankLine
\tcp{Stage 2: change pose after reaching the target body shape}
$\mathcal{B}_{\theta} \leftarrow \textsc{AdaptiveInterpolate}(M,(\boldsymbol{\theta}_s,\boldsymbol{\beta}_e),(\boldsymbol{\theta}_e,\boldsymbol{\beta}_e),\tau)$\;

\BlankLine
\tcp{Concatenate the two paths without duplicating the shared boundary frame}
$\mathcal{B} \leftarrow \mathcal{B}_{\beta} \oplus \mathcal{B}_{\theta}[2{:}]$\;
Align $\mathcal{B}$ to the default body height\;
\Return $\mathcal{B}$\;
\end{algorithm}
\end{minipage}
\end{center}

As illustrated in Algorithm~\ref{alg:smooth_body_sequence}, we first morph the default body to the target shape and then articulate it to the target pose. The operator \textsc{AdaptiveInterpolate} recursively inserts midpoint states when the maximum non-hand body-vertex displacement between two adjacent states exceeds $\tau$. This keeps difficult transitions, such as large torso/hip changes or strong limb rotations, temporally dense while avoiding unnecessary frames in easy segments.

After the smooth body sequence is generated, the cloth is simulated over the sequence with online body updates. At each step, we update the whole-body collision mesh, the body-part meshes used by cloth-reference drag, attachment anchors and normals, and body particle velocities. Algorithm~\ref{alg:dynamic_update} summarizes how the simulator uses the body sequence after the garment has first settled on the default body. During the dynamic stage, body velocities are calculated by moving the body mesh through consecutive states, $\mathbf{v}_t = {\mathbf{x}_t - \mathbf{x}_{t-1}}/{\Delta t}$. The body-part reference meshes and attachment anchors are updated consistently. After the body motion ends, body velocities are reset to prevent residual motion from introducing inertial artifacts into the final cloth state. Unlike static draping pipelines that settle cloth on a fixed body, our simulator keeps cloth coupled to the evolving body throughout the transition, allowing folds, tension, and draping behavior to reflect the final body morphology.

\begin{center}
\begin{minipage}{0.65\linewidth}
\begin{algorithm}[H]
\caption{Dynamic cloth simulation over the body sequence}
\label{alg:dynamic_update}
\DontPrintSemicolon
\SetKwInOut{KwIn}{Input}
\SetKwInOut{KwOut}{Output}

\KwIn{
Current frame $f$; static-detected frame $f_s$; body sequence $\{\mathbf{B}_t\}_{t=0}^{T-1}$; 
body-part sequences $\{\mathbf{B}^{(p)}_t\}_{t=0}^{T-1}$.
}
\KwOut{
Updated body meshes, constraints, and cloth state.
}

\BlankLine
\If{$f_s \leq f < f_s + T$}{
    $t \leftarrow f - f_s$\;
    
    Move the whole-body collision mesh to $\mathbf{B}_t$ and refit it\;

    \ForEach{body part $p$}{
        Move the body-part reference mesh to $\mathbf{B}^{(p)}_t$ and refit it\;
    }

    \If{attachment constraints are enabled}{
        Update attachment points and normals for this body frame\;
    }

    Reset body-collision filters once to enable full cloth--body contact\;
    Run one XPBD cloth-simulation step\;
    Apply optional wearing-style and limb-end constraints\;
}
\Else{
    Set body velocities to zero once and continue post-motion cloth settling\;
}
\end{algorithm}
\end{minipage}
\end{center}

\subsection{Wearing-Style and Limb-End Controls}
\label{sec:supp_upper_lower_tuck}
\label{sec:supp_limb_end_stop}

\begin{figure}[h]
    \centering
    \includegraphics[width=0.9\linewidth]{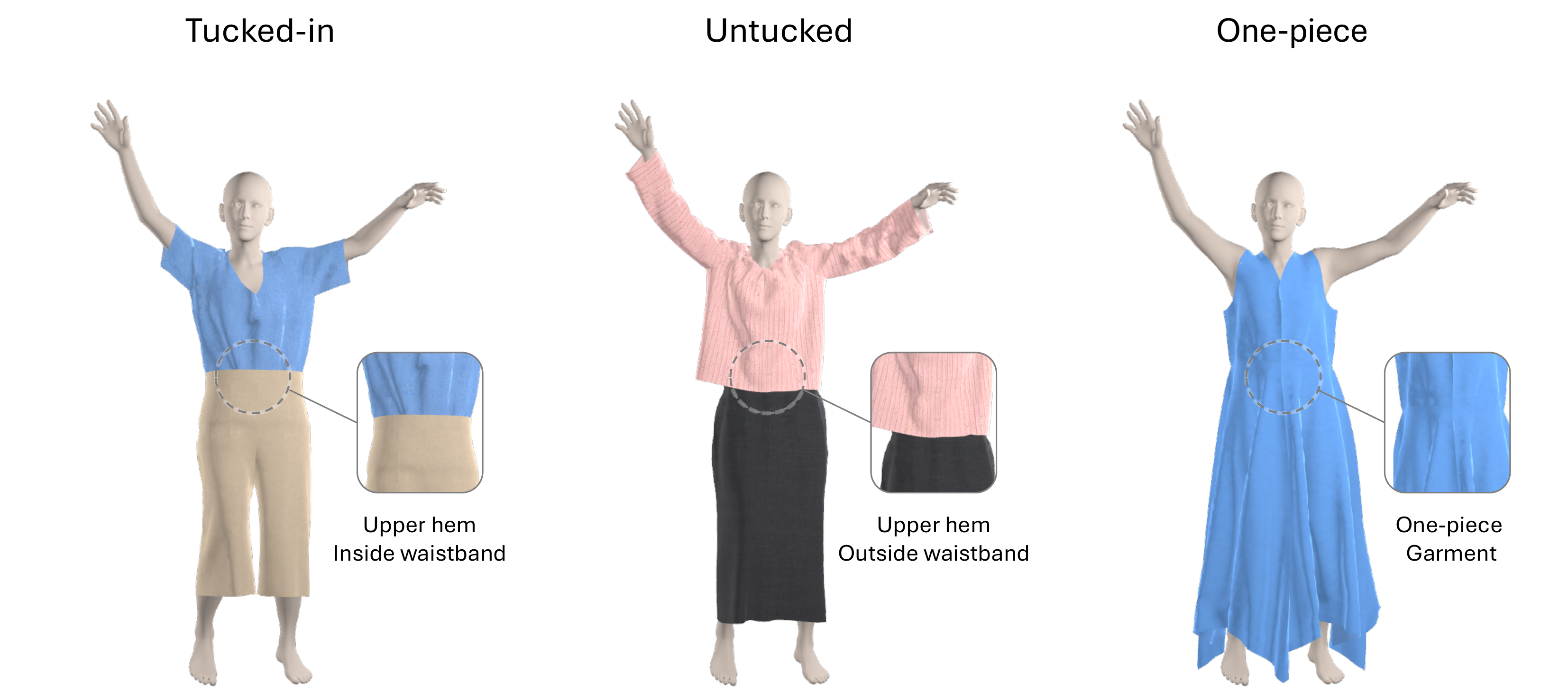}
    \caption{\textbf{Wearing-style control.} The updated simulation engine supports one-piece garments, tucked-in upper--lower outfits, and untucked upper--lower outfits, covering common outfit configurations absent from a one-piece-only GarmentCode setting.}
    \label{fig:supp_wearing_style_comparison}
\end{figure}

\paragraph{Wearing-style control for upper--lower outfits.}
The original GarmentCode setting can represent an outfit as a single stitched garment, which is sufficient for one-piece draping but cannot faithfully model common outfits where the top and bottom are physically separate. It also cannot explicitly distinguish tucked-in and untucked styles, because the relative ordering between the upper hem and lower waistband is not represented. We address this by decomposing compatible patterns into separate upper and lower garment meshes and simulating them jointly on the same dynamic SMPL-X body. The simulator classifies panels into upper-body and lower-body groups, preserves role-specific labels such as upper hem and lower waistband, and uses the shared body sequence and collision field for both garments. For tucked-in outfits, the upper garment is constrained inside the lower waistband region; for untucked outfits, the upper garment is simulated outside the waistband so that it drapes over the lower garment. Figure~\ref{fig:supp_wearing_style_comparison} shows the three supported wearing styles. This extension exposes FitVTON to realistic garment layering and wearing-style-dependent fit behavior during supervision generation.

\begin{figure}[t]
    \centering
    \includegraphics[width=0.8\linewidth]{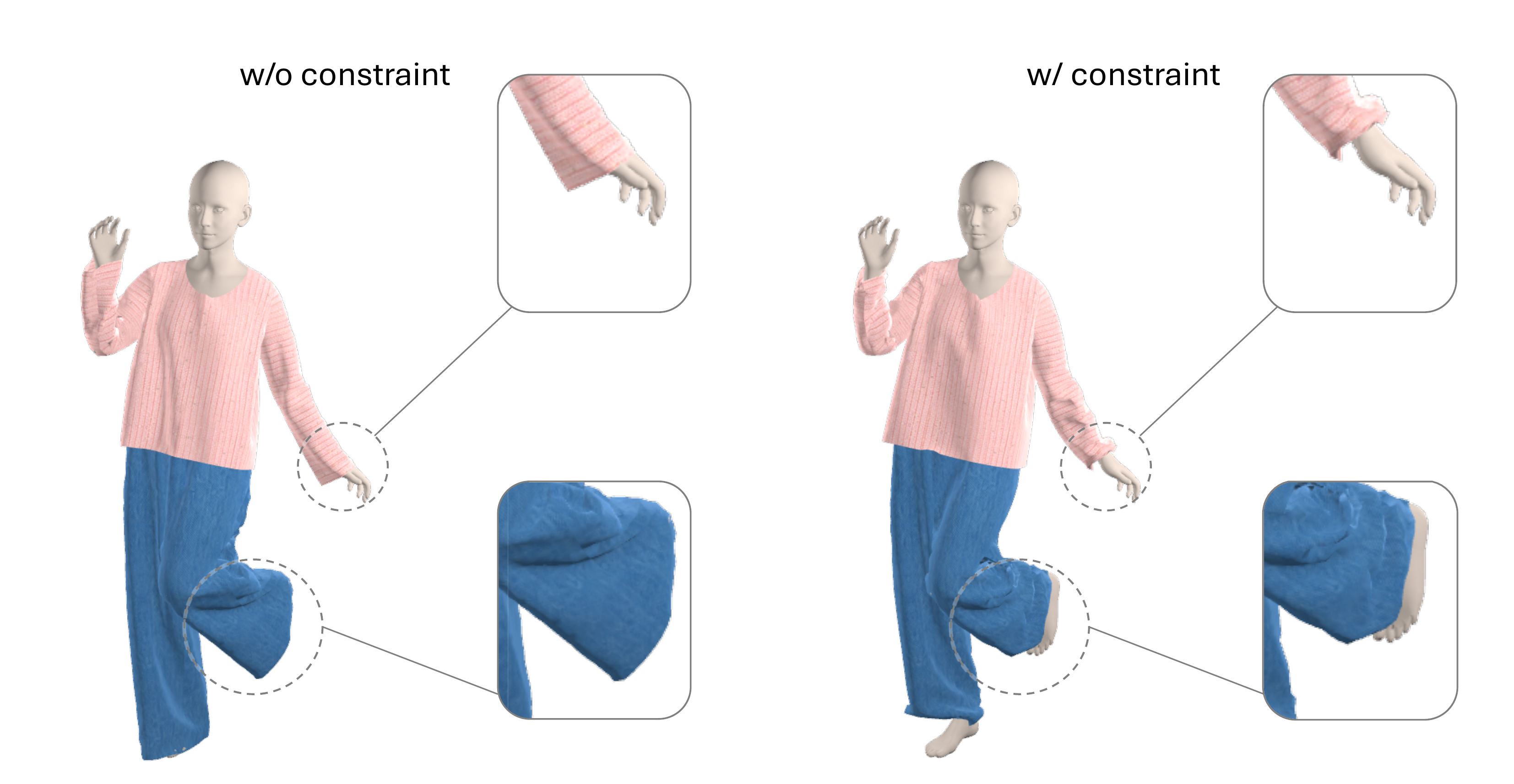}
    \caption{\textbf{Effect of wrist/ankle stop constraints.} Without limb-end stop constraints, sleeve cuffs or pant hems may slide past the hand or foot. With the proposed constraints, garment ends stay near the wrist or ankle while preserving natural local folds.}
    \label{fig:supp_limb_end_stop_comparison}
\end{figure}

\paragraph{Wrist/ankle stop constraints.}
Long sleeves and loose pant legs can drift beyond the wrist or ankle under gravity and body motion, especially in articulated poses where local collision geometry changes rapidly. Such sliding produces implausible supervision, with cuffs passing the hand or hems passing the foot. We introduce a limb-end stopping rule rather than a fixed attachment, so garment ends remain anatomically plausible while still forming wrinkles and folds. For each eligible garment, the simulator selects cloth vertices near sleeve cuffs or pant hems using panel labels, builds a limb axis from proximal to distal body parts, and projects any vertex that crosses the wrist or ankle boundary back to the valid side while damping its velocity. As shown in Figure~\ref{fig:supp_limb_end_stop_comparison}, the constraint stabilizes garment ends without rigidly pinning them to the body.

\subsection{Representative Body-Shape Prototypes}
\label{sec:supp_body_shape_prototypes}

\begin{figure}[t]
    \centering
    \includegraphics[width=0.8\linewidth]{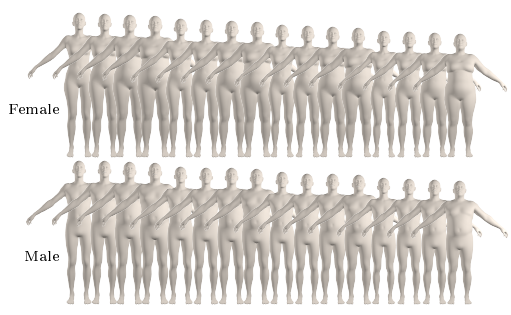}
    \caption{\textbf{Representative female body-shape prototypes.} We use 16 SMPL-X body shapes that span four height categories and four body-size categories, providing a compact set of representative bodies for fitting-aware garment simulation.}
    \label{fig:supp_body_shape_prototypes}
\end{figure}

To cover diverse body shapes with a compact simulation set, we construct 16 representative female and male SMPL-X body prototypes, respectively. Following the anthropometric analysis in SHAPY~\cite{choutas2022accurate}, we organize the prototypes along two interpretable axes: body height and body size. Specifically, we combine four height categories (\textit{short}, \textit{medium short}, \textit{medium tall}, and \textit{tall}) with four body-size categories (\textit{slim}, \textit{medium slim}, \textit{medium plus-size}, and \textit{plus-size}). These prototypes serve as simulation anchors for generating controllable garment-body fit variations. Figure~\ref{fig:supp_body_shape_prototypes} shows 16 variations of female body shapes.

\section{Implementation Details}
\label{sec:supp_implementation_details}

\paragraph{Training setup.}
We perform mask-free try-on training on GarmentCodeVTON using all available triplets. Each training image is randomly cropped or padded to a fixed $4{:}3$ aspect ratio and resized to $1024\times768$, which introduces variation in subject scale and placement. We train for 10 epochs using AdamW with weight decay $1\times10^{-4}$. The learning rate is $1\times10^{-4}$ for the try-on model and $5\times10^{-4}$ for the text encoder. We use a cosine learning-rate schedule with warmup and \texttt{bf16} mixed precision. All experiments use 4 NVIDIA H20 GPUs with a batch size of 2 per GPU.

\paragraph{LoRA adaptation and inference.}
We insert LoRA modules into the Transformer blocks. In stage one, LoRA is applied to self-attention, cross-attention, and their associated feed-forward layers. In stage two, texture-rectification finetuning is performed on fixed VITON and DressCode pseudo triplets for the same number of epochs as stage one. To minimize interference with fitting behavior, stage two updates only self-attention and the corresponding feed-forward layers. For all LoRA modules, the rank is 32 and the scaling factor $\alpha$ is 64. At inference time, we linearly blend the stage-one and stage-two LoRA parameters with mixing ratios of 0.8 and 0.2, respectively, and use 30 sampling steps with classifier-free guidance~\cite{ho2022classifier} scale 1.0.

\section{VLM Evaluator Details}
\label{sec:supp_vlm_evaluator}

\begin{table*}[t]
\centering
\caption{
Repeated-run stability of the VLM evaluator. We report the sample-level standard deviation of the averaged score, dimension-wise score standard deviations, and semantic similarity of textual explanations. For standard deviation, lower is better; for explanation similarity, higher is better.
}
\label{tab:vlm_repeat_consistency}
\small
\setlength{\tabcolsep}{4pt}
\renewcommand{\arraystretch}{1.08}
\begin{tabular}{lcccccccccc}
\toprule
Category
& $N$
& Avg Std$\downarrow$
& \multicolumn{4}{c}{Score Std$\downarrow$}
& \multicolumn{4}{c}{Explanation Sim.$\uparrow$} \\
\cmidrule(lr){4-7}\cmidrule(lr){8-11}
& & & GB & T/L & SC & LF & GB & T/L & SC & LF \\
\midrule
Upper & 200 & 0.074 & 0.020 & 0.084 & 0.000 & 0.242 & 0.934 & 0.927 & 0.911 & 0.886 \\
Lower & 200 & 0.066 & 0.024 & 0.118 & 0.129 & 0.138 & 0.926 & 0.914 & 0.907 & 0.895 \\
Dress & 100 & 0.071 & 0.080 & 0.089 & 0.169 & 0.267 & 0.915 & 0.898 & 0.894 & 0.868 \\
\bottomrule
\end{tabular}
\end{table*}

\subsection{Repeated-Run Consistency}
\label{sec:supp_vlm_consistency}

We assess the robustness of the VLM-based fitting evaluator by evaluating the same image pairs five times under a fixed protocol using GPT-5.2, temperature= 0, and random seed 42. The subset contains 500 \textit{FittingEffect3K} samples: 200 upper-body, 200 lower-body, and 100 dress samples. For each sample, we record four dimension scores, compute their standard deviation across runs, and also compute the standard deviation of the averaged four-dimension score.

% \begin{figure}[t]
%     \centering
%     \includegraphics[width=1\linewidth]{Figure12.pdf}
%     \caption{Sample-level averaged score variability across repeated runs. For each sample, the averaged score is the mean of the four dimension scores. }
%     \label{fig:vlm_repeat_sample_std}
% \end{figure}

Table~\ref{tab:vlm_repeat_consistency} reports mean sample-wise standard deviations ranging from 0.0658 to 0.0738, indicating limited variation in the four-dimension average fitting score across repeated runs. The dimension-wise score standard deviations are also stable overall. GB and T/L are generally more stable, while LF shows larger variance for Upper and Dress garments because local artifacts depend on finer and more localized visual cues.

\subsection{Explanation Consistency}
\label{sec:supp_vlm_explanation_consistency}

We also evaluate whether the textual explanations are consistent across repeated runs. For each evaluation dimension, we compute average pairwise semantic similarity between explanations using the gte-large-en-v1.5 embedding model on the same 500-sample subset.

Table~\ref{tab:vlm_repeat_consistency} shows high explanation similarity across categories and dimensions. Upper and Lower similarities are all above 0.88 and mostly above 0.90; Dress remains high, ranging from 0.8684 to 0.9154. LF is slightly less consistent than GB and T/L, matching the score-variance trend and reflecting the finer visual granularity required to judge local fit artifacts.

\subsection{Prompt Design}
\label{sec:supp_vlm_prompt}

\begin{table}[t]
\centering
\caption{
Category-specific fitting focuses used in the VLM evaluator prompts.
}
\label{tab:vlm_prompt_dimensions}
\scriptsize
\setlength{\tabcolsep}{3pt}
\renewcommand{\arraystretch}{1.04}
\begin{tabular}{p{0.18\linewidth} p{0.36\linewidth} p{0.38\linewidth}}
\toprule
Category & Alignment Focus & Tightness / Silhouette / Artifacts Focus \\
\midrule
Upper-body
& Shoulder line, neckline, armhole, sleeve length, top length
& Chest/bust, shoulders, armpits, upper arms, torso volume, neckline, sleeves \\

Lower-body
& Waist position, rise/crotch placement, hip line, pant length
& Waist, hips, crotch, thighs, knees, calves, leg volume, inner thighs, hem \\

Dress / Skirt
& Shoulder/upper alignment if applicable, waistline, hip line, hem length
& Bust if applicable, waist, hips, upper thighs, full-body silhouette, waist--hip transition, side seams, hem \\
\bottomrule
\end{tabular}
\end{table}

We design a structured prompting protocol for fit-oriented VLM evaluation. 
The system prompt defines the evaluator role, the scoring rubric, and the required JSON output format. 
It explicitly instructs the evaluator to ignore image-level nuisance factors, such as lighting, exposure, background, camera perspective, framing, and apparent person scale, and to focus instead on garment-to-body placement, body-region-relative proportions, silhouette consistency, and local fit behavior.

For category-specific evaluation, we use three user prompts for upper-body, lower-body, and dress/skirt garments. 
All prompts follow the same evaluation scaffold, including four dimensions: garment--body alignment, tightness/looseness consistency, silhouette consistency, and local fit artifacts. 
The concrete body regions and output keys are adapted to each garment category, as summarized in Table~\ref{tab:vlm_prompt_dimensions}. 
The full system prompt is provided below.

\begin{promptlisting}{System Prompt}
SYSTEM_PROMPT = """You are a professional garment fitting evaluator with expertise in apparel design and human body fitting.

Your task is to compare garment fitting quality between two images of the same garment worn by the same person or people with the same body shape.

One image shows the real person wearing the garment (ground truth).
The other image shows a virtual try-on result.

You must treat the real-wearing image as the ground truth reference.

Do NOT judge visual aesthetics, image realism, fashion style, attractiveness, or global image composition.
Focus strictly on fitting accuracy by comparing where the garment sits on the body, how it aligns with specific body regions, and whether its proportions, silhouette, and local fit behavior are consistent with Image A.
Ignore nuisance differences caused by lighting, exposure, white balance, shadows, background, camera perspective, camera distance, focal length, framing, or the person's apparent image scale.

You must follow these output rules strictly.

Scoring rubric (use INTEGER scores 1--5 only):
- 5: Virtually identical fit to Image A; only negligible differences.
- 4: Very close; minor, localized differences that do not change the perceived fit behavior.
- 3: Acceptable but clearly different; multiple noticeable differences in fit behavior, silhouette, or alignment.
- 2: Poor match; major discrepancies that change key fit behavior or proportions.
- 1: Very poor; the fit behavior is largely incorrect or inconsistent with Image A.

For each required dimension:
- Provide `score` as an integer from 1 to 5.
- Provide a brief `explanation` grounded in visible fitting differences.

Finally:
- Provide a short `summary` of the main category-relevant fitting discrepancies.

Output ONLY a valid JSON object. Do NOT include any extra text, markdown, or code fences.
"""
\end{promptlisting}

\section{Supplementary Visualization Results}
\label{sec:supp_visualization_results}
\label{sec:supp_prompt_fitting}
\label{sec:supp_ablation_results}

This final section provides visual evidence that complements the main-paper results. We first analyze whether stronger text prompts alone can induce reliable fitting behavior, then show additional qualitative comparisons with baseline methods, and finally present supplementary ablation visualizations.

\subsection{Prompt-Based Fitting Analysis}

\begin{figure*}[t]
    \centering
    \includegraphics[width=0.95\linewidth]{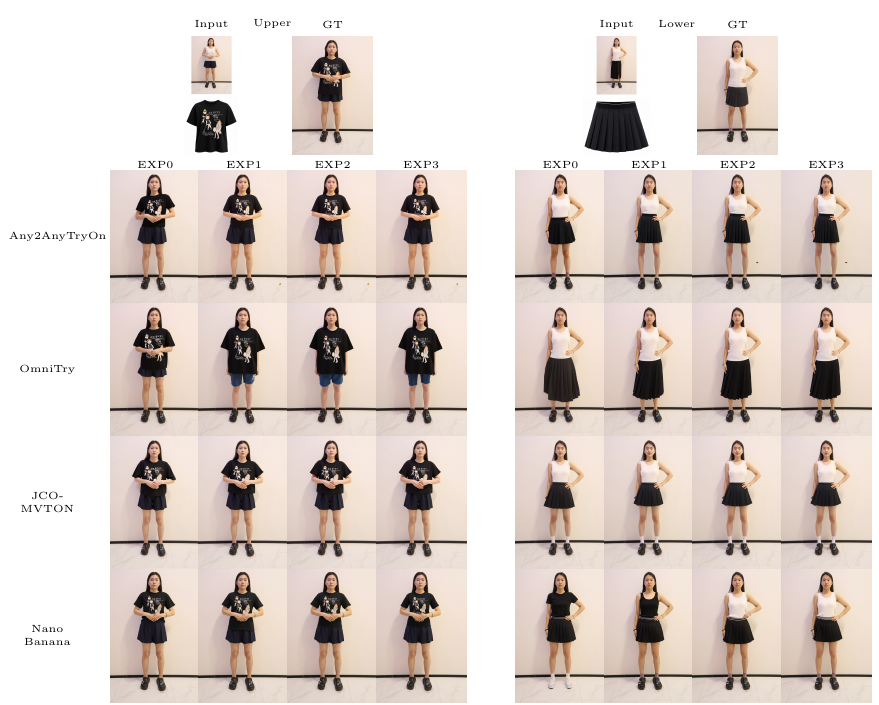}
    \caption{Supplementary qualitative results of prompt-based fitting tests on upper-body (left) and lower-body (right) garments. EXP0--EXP3 denote the base, body-attribute, metric-attribute, and combined-attribute prompts, respectively. Stronger prompting does not consistently improve garment--body fitting.}
    \label{fig:prompt_fitting_test}
\end{figure*}

We further investigate whether current virtual try-on models and general-purpose image generation models can acquire fitting awareness through stronger textual conditioning. Specifically, we evaluate four models, including specialized virtual try-on models and the commercial image generation model Nano Banana, on selected person-garment pairs from \textit{FittingEffect3K}. For each pair, we keep the input person image and garment image fixed, and vary only the textual prompt.

We compare four prompt settings. The \emph{base prompt} uses the original try-on instruction without additional fitting information. The \emph{body-attribute prompt} augments the base prompt with high-level descriptions of the person and garment, such as body shape, height category, and garment size. The \emph{metric-attribute prompt} provides more explicit measurements, including person height, body weight, and garment length. The \emph{combined-attribute prompt} includes both body-level and metric-level descriptions.

As shown in Figure~\ref{fig:prompt_fitting_test}, stronger textual conditioning does not lead to reliable fitting behavior.Across different models and garment categories, the generated images may remain visually plausible, but the garment-person relationship is not physically grounded. The models often alter the apparent garment length, body proportion, or clothing silhouette in ways that are inconsistent with the provided attributes. Moreover, adding more detailed prompts does not consistently improve results over the base prompt; in some cases, it even introduces more severe scale, length, or deformation artifacts.

These results suggest that current models can perceive and react to body- or garment-related textual cues, but such reactions are closer to learned visual adjustment than true garment-body fitting. In other words, they synthesize plausible try-on images from learned visual priors, rather than explicitly reasoning about whether a specific garment size, length, or silhouette fits a specific human body. Therefore, prompt engineering alone is insufficient to endow current try-on or commercial image generation models with reliable fitting capability, highlighting the need for explicit fitting-aware modeling.

\subsection{Additional Comparison and Ablation Visualizations}

\begin{figure*}[t]
  \centering

  \begin{minipage}[t]{0.54\textwidth}
    \centering
    \includegraphics[width=\linewidth]{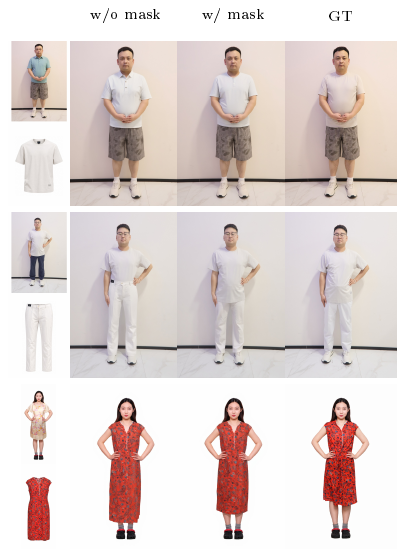}
  \end{minipage}\hfill
  \begin{minipage}[t]{0.4\textwidth}
    \centering
    \includegraphics[width=\linewidth]{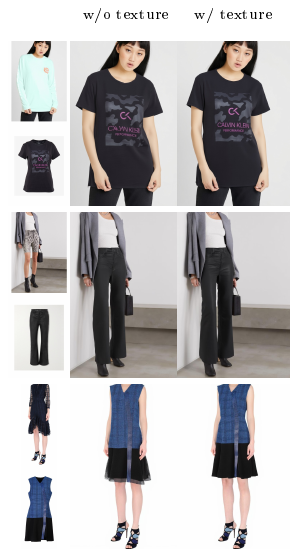}
  \end{minipage}
  \caption{
  Supplementary qualitative ablation results. 
  (a) Dual-branch mask supervision improves garment localization and fitting consistency. 
  (b) Texture rectification improves high-frequency garment details while preserving the try-on structure.
  }
  \label{fig:supp_ablation}
\end{figure*}

\begin{figure*}[t]
  \centering
  \includegraphics[width=\textwidth]{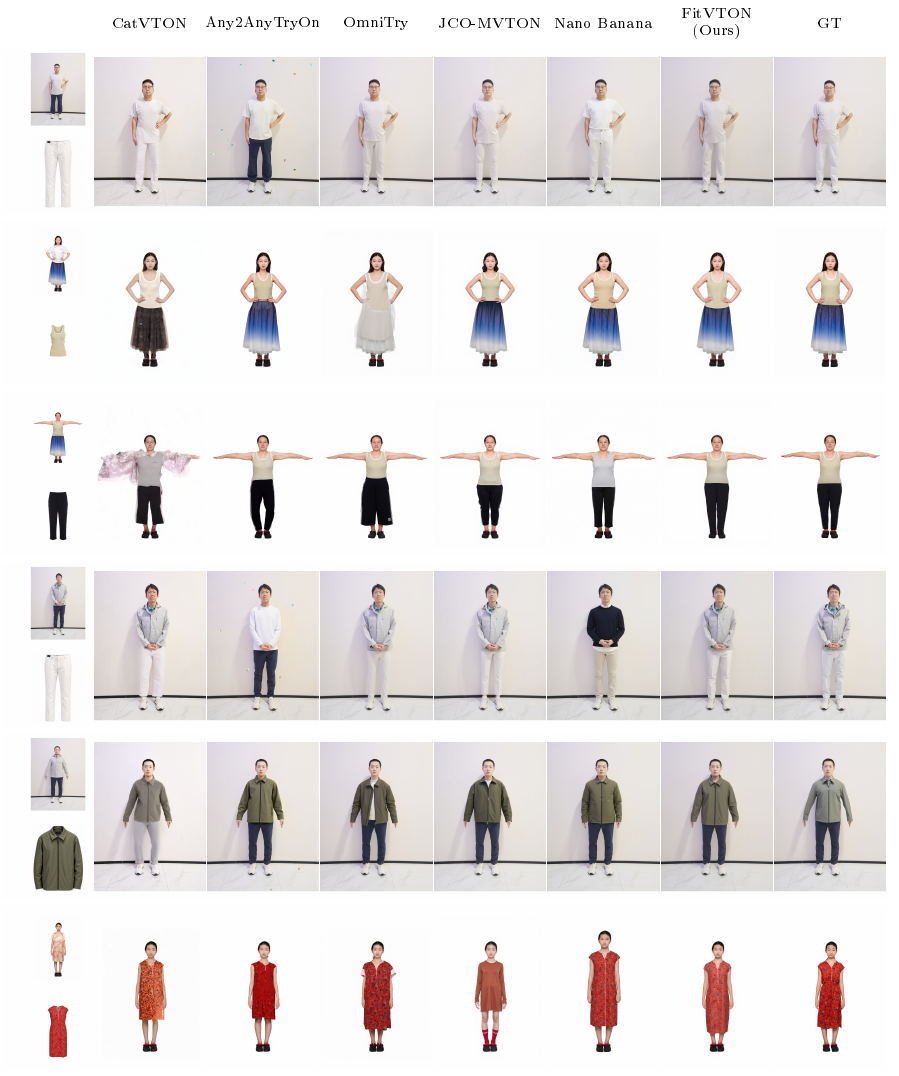}
  \caption{\textbf{Additional fit-oriented comparison with baseline methods.} We provide more virtual try-on examples comparing FitVTON with representative baseline methods. These examples complement the main-paper results by showing garment appearance preservation, body-structure consistency, and fitting behavior across additional samples.}
  \label{fig:supp_fit_comparison}
\end{figure*}

\begin{figure*}[t]
  \centering
  \includegraphics[width=0.83\textwidth]{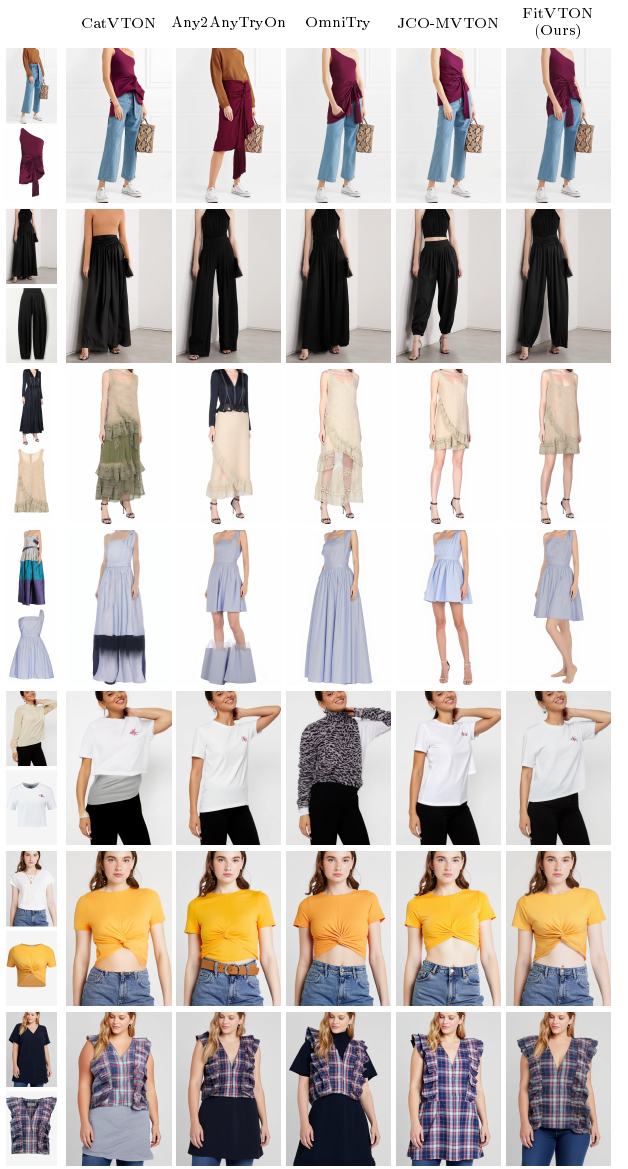}
  \caption{\textbf{Additional qualitative comparison with baseline methods.} We provide more virtual try-on examples comparing FitVTON with representative baseline methods. These examples complement the main-paper results by showing garment appearance preservation, body-structure consistency, and texture realism.}
  \label{fig:supp_quality_comparison}
\end{figure*}

We provide additional qualitative examples that complement the main-paper visual results. Figure~\ref{fig:supp_fit_comparison} and Figure~\ref{fig:supp_quality_comparison} compares FitVTON with representative baseline methods on more virtual try-on samples, while Figure~\ref{fig:supp_ablation} shows supplementary ablation cases for the dual-branch mask supervision and texture-rectification stage.

